# What a Nerd! Beating Students and Vector Cosine in the ESL and TOEFL Datasets

**Enrico Santus***, **Tin-Shing Chiu***, **Qin Lu***, **Alessandro Lenci**§, **Chu-Ren Huang***

\* The Hong Kong Polytechnic University, Hong Kong
esantus@gmail.com, cstschiu@comp.polyu.edu.hk, {qin.lu, churen.huang}@polyu.edu.hk
§ University of Pisa, Italy
alessandro.lenci@unipi.it

**Abstract**

In this paper, we claim that *Vector Cosine* – which is generally considered one of the most efficient unsupervised measures for identifying word similarity in *Vector Space Models* – can be outperformed by a completely unsupervised measure that evaluates the extent of the intersection among the most associated contexts of two target words, weighting such intersection according to the rank of the shared contexts in the dependency ranked lists. This claim comes from the hypothesis that similar words do not simply occur in similar contexts, but they share a larger portion of their most relevant contexts compared to other related words. To prove it, we describe and evaluate *APSyn*, a variant of *Average Precision* that – independently of the adopted parameters – outperforms the *Vector Cosine* and the *co-occurrence* on the ESL and TOEFL test sets. In the best setting, *APSyn* reaches 0.73 accuracy on the ESL dataset and 0.70 accuracy in the TOEFL dataset, beating therefore the non-English US college applicants (whose average, as reported in the literature, is 64.50%) and several state-of-the-art approaches.

**Keywords:** Vector Space Models, VSMs, Distributional Semantic Models, DSMs, Semantic Relations, Words Similarity

## 1. Introduction

Word similarity detection plays an important role in Natural Language Processing (NLP), as it is the backbone of several applications, such as *Paraphrasing*, *Query Expansion*, *Word Sense Disambiguation*, *Automatic Thesauri Creation*, and so on (Terra and Clarke, 2003).

Several approaches have been proposed to measure word similarity (Terra and Clarke, 2003; Jarmasz and Szpakowicz, 2003; Mikolov et al., 2013; Levy et al., 2015; Santus et al. 2016a). Some of them rely on knowledge resources (such as lexicons or semantic networks), while others are corpus-based.

The latter approaches generally exploit the *Distributional Hypothesis*, according to which words that occur in similar contexts also have similar meanings (Harris, 1954). Although these approaches extract statistics from large corpora, they vary in the way they define what has to be considered context (i.e. lexical context, syntactic context, documents, etc.), how the association with such context is measured (e.g. frequency of co-occurrence, association measures like Pointwise Mutual Information, etc.), and how the association with the contexts is used to identify the similarity (Terra and Clarke, 2003; Hearst, 1992; Santus et al., 2014a; Santus et al., 2014b; Santus et al., 2016a).

A common way to represent word meaning in NLP is by using vectors to encode the association between the target words and their contexts. The resulting vector space is generally referred as *Vector Space Model* (VSM) or, more specifically, as *Distributional Semantic Model* (DSM). In such vector space, word similarity can be calculated by using the *Vector Cosine*, which measures the angle between the vectors (Turney and Pantel, 2010). Other measures – such as *Manhattan Distance*, *Dice's Coefficient*, *Euclidean Distance*, *Jaccard Similarity* and *Matching Coefficient* – can be used to calculate the distance between the vectors (Gomaa and Fahmy, 2013), but the *Vector Cosine* is generally considered to be the optimal choice (Bullinaria and Levy 2007).

Another common way to represent word meaning is using word embeddings, which are vector-space word representations that are implicitly learned by the input-layer weights of neural networks. These models have shown a strong ability to capture synonymy and analogies (such as in the famous "King - Man + Woman = Queen" example, where Mikolov et al. (2013) subtract the vector of "Man" from the one of "King", and then add the vector of "Woman", obtaining a very similar vector to the one of "Queen"), even though Levy et al. (2015) have claimed that traditional count-based DSMs can achieve the same results if their hyperparameters are properly optimized.

A well-known problem with the distributional approaches is that they rely on a very loose definition of similarity. In fact, vectors have as nearest neighbours not only synonyms, but also hypernyms, co-hyponyms, antonyms, as well as a wide range of other semantically related items (Santus et al., 2015).

For this reason, several datasets have been proposed by the NLP community to test distributional similarity measures. Among the most common ones, there are the *English as a Second Language*[1] dataset (ESL; Turney, 2001) and the *Test of English as Foreign Language*[2] (TOEFL; Landauer and Dumais, 1997). The former consists of 50 multiple-choice synonym questions, with 4 choices each, while the latter consists of 80 multiple-choice synonym questions, with 4 choices each.

---
[1] For the state-of-the-art on the ESL, see:

In this paper, we describe and evaluate *APSyn*, a completely unsupervised measure that calculates the extent of the intersection among the *N* most related contexts of two target words, weighting such intersection according to the rank of the shared contexts in a mutual dependency ranked list.

In our experiments, *APSyn* outperforms the *Vector Cosine* and the co-occurrence frequency, reaching 0.73 accuracy on the ESL dataset and 0.70 accuracy in the TOEFL dataset, beating therefore the non-English US college applicants (whose average, as reported in the literature, is 64.50%) and several state-of-the-art approaches.

## 2. Background

Word similarity measures play a fundamental role in tasks such as *Information Retrieval* (IR), *Text Classification* (TC), *Text Summarization* (TS), *Question Answering* (QA), *Sentiment Analysis* (SA), and so on (Terra and Clarke, 2003; Tungthamthiti et al., 2015). They can be either knowledge-based or corpus-based (Gomaa and Fahmy, 2013). The former rely on lexicons or semantic networks, such as WordNet (Fellbaum, 1998), measuring the distance between the nodes in the network. The latter, instead, compute the similarity between words relying on statistical information about their distributions in large corpora (Church and Hanks, 1990).

Knowledge based approaches generally exploit hand-crafted resources. While being hand-crafted ensures high quality, it also entails arbitrariness and high development and update costs. This is the main reason why these resources are known for their limited coverage (Santus et al., 2015b). Such limitation has often prompted researchers to pursue hybrid approaches (Turney, 2001).

A key assumption of corpus-based approaches is that similarity between words can be measured by looking at words co-occurrences. In particular, following the *Distributional Hypothesis* (Harris, 1954; Firth, 1957), these methods assume that words occurring in similar contexts are also similar. These methods mainly vary according to two dimensions: i) how they define the contexts (e.g. document, paragraph, sentence, fixed-size window, etc.); and ii) how they measure whether the targets occur in similar contexts (e.g. weighted occurrence frequency, extent of the intersection, etc.). These models are generally referred as traditional count-based DSMs.

A well-known traditional count-based DSM applied to synonymy identification is the *Latent Semantic Analysis*

---

(LSA; Landauer and Dumais, 1997). This system was tested on the 80 multiple-choice synonym questions of the TOEFL, achieving an accuracy of 64.38%, which is very close to the reported average of non-English US college applicant (i.e. 64.50%).

Another interesting way to learn words statistics is generally referred as *word embeddings* and is discussed in Mikolov et al. (2013). The authors report that when a neural network language model is trained, it is not only the model that is obtained, but also distributed words representations, which can be eventually used for other goals, such as in Collobert and Weston (2008). In their paper, Mikolov et al. (2013) show that these words representations capture both syntactic and semantic regularities, performing particularly well in word similarity identification and analogies.

While such models have obtained an enthusiastic reception, with a consequent boost of papers using word embeddings, Levy et al. (2015) have demonstrated that similar results can be also obtained with optimized traditional count-based DSMs.

### 2.1 Distance Measures

Independently from the approach that is used to learn words statistics, corpus-based approaches represent word meanings as vectors in *vector spaces*, generally called *semantic spaces*. In such semantic spaces, words similarity can be measured as the proximity between vectors. Several measures have been adopted to this scope. In the following rows, we briefly describe some of them, while defining the *Vector Cosine*, which is generally considered the most efficient one.

*Manhattan Distance* (*L1*) can be defined as the sum of the differences of the dimensions. *Euclidean Distance* (*L2*) is the square root of the sum of the squared differences of the dimensions. *Dice's Coefficient* is instead twice the number of common dimensions, divided by the total number of dimensions in the two vectors. *Jaccard Similarity* is defined as the number of shared dimensions divided by the number of unique dimensions in both the vectors. *Matching Coefficient* is the number of dimensions different from zero in both the vectors. *Vector Cosine* looks instead at the normalized correlation between the dimensions of two words, $w_1$ and $w_2$, and is described by the following equation:

$$\cos(w_1, w_2) = \frac{\sum_{i=1}^{n} f_{1i} \times f_{2i}}{\sqrt{\sum (f_{1i})^2} \times \sqrt{\sum (f_{2i})^2}}$$

where $f_{xi}$ is the *i*-th dimension in the vector *x*.

This measure has been extensively used to identify word similarity in vector spaces becoming a sort of *de facto* standard in distributional semantics (Landauer and Dumais, 1997; Jarmasz and Szpakowicz, 2003; Mikolov et al., 2013; Levy et al., 2015).

## 2.2 State-of-the-art in the ESL and TOEFL

After its first use in Landauer and Dumais (1997), the TOEFL dataset became one of the most common benchmarks for vector space models testing: Karlgren and Sahlgren (2001), Pado and Lapata (2007), Turney (2001), Turney (2008), Terra and Clarke (2003), Bullinaria and Levy (2007), Matveeva et al. (2005), Dobó and Csirik (2013) and Rapp (2003). Bullinaria and Levy (2012) even achieved 100% accuracy on this dataset. In their paper, the authors extensively analyze numerous parameters, including the influence of corpus size, window size, stop-lists, stemming and Singular Values Decomposition (SVD), until they find a perfectly optimized model. After achieving perfect precision on the TOEFL, the authors acknowledge that while these results are impressive for the benchmark, they can hardly be generalized to new tasks.

Few years after the introduction of TOEFL as a benchmark, Turney (2001) proposed the ESL. These 50 multiple-choice synonym questions are provided in a sentence context, to facilitate sense disambiguation. ESL has soon become a very popular benchmark on which several models have been evaluated. The best reported corpus-based approaches in this benchmark were those of Turney (2001), Terra and Clarke (2003) and Jarmasz and Szpakowicz (2003). The latter achieving the best result of 82% accuracy.

## 3. Method

Given a traditional count-based DSM, where every word is represented as a vector of weighted associations between such word and contexts, we can re-think the *Distributional Hypothesis* (Harris, 1954) by hypothesizing that similar words not only occur in similar contexts, but – more specifically – they share a larger number of their most associated contexts, compared to less similar ones.

A way to test this hypothesis is by: i) measuring the extent of the intersection among the $N$ most related contexts of two target words, and ii) weighting such intersection according to the rank of the shared contexts in the dependency ranked lists. This can be done in several steps. First of all, for every target word we rank the contexts according to the *Positive Pointwise Mutual Information* (PPMI; Levy et al., 2015), which is defined as follows:

$$PMI(w,c) = \log\left(\frac{P(w,c)}{P(w) \times P(c)}\right) = \log\left(\frac{|w,c| \times D}{|w| \times |c|}\right)$$

$$PPMI(w,c) = \max(PMI(w,c), 0)$$

were $w$ is the target word, $c$ is the given context, $P(w,c)$ is the probability of co-occurrence and $D$ is the collection of observed word-context pairs.

Once the contexts are ranked according to their PPMI, for every target word we pick the top $N$ contexts and we intersect them. At this point, for each shared context, we add one divided by the average rank of the shared context in the PPMI-ranked contexts lists. We formalize this as the *APSyn* similarity measure:

$$APSyn(w_1, w_2) = \sum_{f \in N(F_1) \cap N(F_2)} \frac{1}{(rank_1(f) + rank_2(f))/2}$$

For every feature $f$ included in the intersection between the top $N$ features of $w_1$, $N(F_1)$, and $w_2$, $N(F_2)$, *APSyn* will add 1 divided by the average rank of the feature, among the top PPMI ranked features of $w_1$, $rank_1(f_1)$, and $w_2$, $rank_2(f_2)$.

The choice of the weighting function is a parameter of APsyn. In previous experiments, published in Santus et al. (2016a), we used *Local Mutual Information* (LMI; Evert, 2005) to rank the contexts, instead of using PPMI. However, the LMI-ranked *APSyn* obtained worse results than those reported in the current paper. Such results were nonetheless still outperforming the *Vector Cosine* and the co-occurrence frequency. In section 9, we will comment on the differences.

## 4. Evaluation

In the following paragraphs we describe our DSMs, the test sets and the task.

### 4.1 Distributional Semantic Model

We use several window-based DSMs, recording word co-occurrences within the $K$ nearest content words to the left and right of each target, where K has the following values: 2, 3, 5 and 10. Co-occurrences are extracted from a combination of ukWaC and WaCkypedia corpora (around 2.7 billion words) for content words – namely adjectives, nouns and verbs – occurring over 1,000 times, and are weighted with PPMI. The model consists of 28,870 word vectors, each of which with 28.870 dimensions.

### 4.2 Test Sets

In order to evaluate the proposed measure, we use both the ESL (Turney, 2001) and TOEFL (Landauer and Dumais, 1997) datasets. The former consists of 50 questions, while the latter of 80 questions. The ESL sentences were not used in our experiments. An example of ESL question is the following:

- "An underground [**passage**] connected the house to the garage."
    a. **Hallway**
    b. Ticket
    c. Entrance
    d. Room

For both datasets, we have turned each question in four pairs, each of which containing the *problem word* and apossible answer. Unfortunately, we do not have a full coverage of the datasets, because our model was built for content words with frequency over 1000, Parts-Of-Speech-tagged either as adjectives, nouns or verbs. In the ESL test set, 4 out of 50 questions were excluded because the correct answers were not present in the DSM. In the TOEFL test set, 20 out of 80 questions were excluded for the same reason. Few questions, moreover, have one missing choice. In order to keep them for the evaluation, in case of correct answer, the score is increased of $0.25 * |choices\ in\ DSM|$ (where, 1 is added only if all four choices are in the DSM).

### 4.3 Task

We have assigned *APSyn* scores to all the pairs, and then – for every problem word – we have sorted the possible choices in a decreasing order. We considered positive every problem word having the correct answer on top, negative all the others.

### 5. Results

In Table 1, we report the results of *APSyn* and the baselines in the ESL test.

| APSyn | Win 2 | Win 3 | Win 5 | Win 10 |
|---|---|---|---|---|
| N=100 | **0.73** | 0.69 | 0.67 | 0.62 |
| N=200 | 0.71 | 0.69 | 0.67 | 0.62 |
| N=300 | 0.71 | 0.69 | 0.69 | 0.62 |
| N=400 | 0.71 | 0.69 | 0.67 | 0.62 |
| N=500 | 0.71 | 0.69 | 0.67 | 0.62 |
| N=600 | 0.71 | 0.69 | 0.65 | 0.62 |
| N=700 | 0.71 | 0.69 | 0.65 | 0.67 |
| N=800 | 0.71 | 0.69 | 0.67 | 0.67 |
| N=900 | 0.69 | 0.65 | 0.67 | 0.65 |
| N=1000 | 0.67 | 0.62 | 0.65 | 0.60 |
| Baselines | | | | |
| Cosine | 0.46 | 0.46 | **0.48** | **0.48** |
| Co-occ | 0.43 | 0.41 | 0.39 | 0.35 |

Table 1. Accuracy in the ESL test set for *APSyn* (100<*N*<1000), *Vector Cosine* and co-occurrence, in window 2, 3, 5 and 10 DSMs.

As it can be seen from Table 1 and from Figure 1, *APSyn* always outperforms the baselines.
The window size and *N* have a certain impact on the performance of *APSyn*. The former parameter has an impact also on the baselines (*Vector Cosine* seems to perform slightly better for larger windows, while the co-occurrence frequency seems to prefer smaller ones). Our measure, in particular, seems to perform better on smaller windows and for *N* close to 100, while its performance slightly drops for *N* close to 1000.
A possible reason for such drop may be that if too many contexts are considered, some rumor is added. This happens because with larger values of *N*, *APSyn* is forced to consider less important contexts of the targets.
In Table 2, we report the scores for *APSyn* and the baselines in the evaluation on the TOEFL test set (see also Figure 2). *APSyn* outperforms the baselines, especially when the window size and *N* are small. Interestingly, the *Vector Cosine* prefers a smaller window in this dataset (this is actually what we would have expected also for the ESL, as smaller windows are known to better capture paradigmatic similarity).

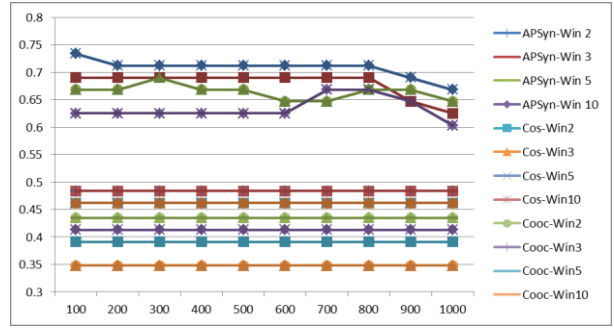

Figure 1. Accuracy in the ESL test set for *APSyn* (100<*N*<1000), *Vector Cosine* and co-occurrence, in window 2, 3, 5 and 10 DSMs.

| APSyn | Win 2 | Win 3 | Win 5 | Win 10 |
|---|---|---|---|---|
| N=100 | 0.68 | 0.67 | 0.62 | 0.59 |
| N=200 | 0.68 | 0.65 | 0.62 | 0.61 |
| N=300 | **0.70** | 0.67 | 0.62 | 0.61 |
| N=400 | **0.70** | 0.67 | 0.62 | 0.61 |
| N=500 | **0.70** | 0.67 | 0.62 | 0.61 |
| N=600 | 0.68 | 0.67 | 0.62 | 0.61 |
| N=700 | 0.68 | 0.65 | 0.62 | 0.62 |
| N=800 | 0.67 | 0.67 | 0.62 | 0.62 |
| N=900 | 0.67 | 0.65 | 0.66 | 0.62 |
| N=1000 | 0.65 | 0.67 | 0.68 | 0.59 |
| Baselines | | | | |
| Cosine | **0.58** | 0.53 | 0.50 | 0.46 |
| Co-occ | 0.45 | 0.41 | 0.44 | 0.45 |

Table 2. Accuracy in the TOEFL test set for *APSyn* (100<*N*<1000), *Vector Cosine* and co-occurrence, in window 2, 3, 5 and 10 DSMs.

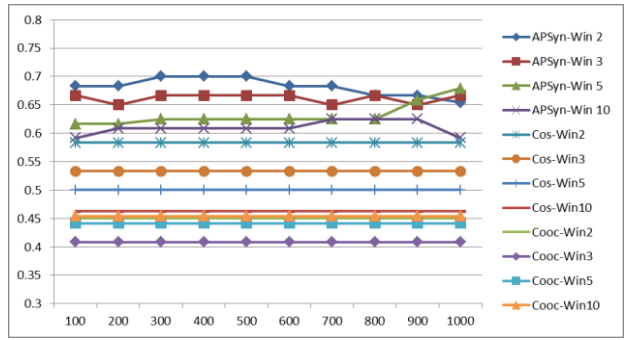

Figure 2. Results in the TOEFL test set for *APSyn* (with *N* ranging from 100 to 1000, with increments of 100), *Vector Cosine* and co-occurrence, in different DSMs.

While considering the results, it must be kept in mind that those that may look as high variances, could be actually be very small ones, given the small size of the test sets. Guessing one more question, for example, would have a large impact on the accuracy.

## 6. Error Analysis

In this section we briefly analyse the best and worst performance of *APSyn* in the ESL dataset.

From the previous section, we have seen that *APSyn* performs best with a window 2 DSM and *N*=100. In Table 3, we report the non-identified synonyms (with their Parts of Speech: n=noun, v=verb and j=adjective), with their actual rank in the question pairs. As it can be seen, most of the non-identified synonyms are ranked second according to *APSyn*, and they are generally placed after a word that – at least in certain contexts – has a very similar meaning to the problem word (e.g. *grind-slice*, *limb-trunk*, *passage-entrance*, *refer-call*, *scrape-slice*, *steep-rugged*, *twist-curl*). Unfortunately, we have not used the contextual sentences to disambiguate the problem words.

| Problem word | 1st | 2nd | 3rd | 4th |
|---|---|---|---|---|
| grind-v | slice-v | **rub-v** | tap-v | hit-v |
| harvest-n | stem-n | **intake-n** | lump-n | split-n |
| limb-n | trunk-n | **branch-n** | twig-n | bark-n |
| passage-n | entrance-n | **hallway-n** | room-n | ticket-n |
| paste-n | syrup-n | **dough-n** | jelly-n | block-n |
| refer-v | call-v | explain-v | **direct-v** | carry-v |
| scrape-v | slice-v | **grate-v** | chop-v | mince-v |
| steep-j | rugged-j | bare-j | **sheer-j** | --- |
| substance-n | level-n | **thing-n** | posture-n | score-n |
| twist-v | curl-v | clip-v | fasten-v | **intertwine-v** |
| yield-v | scorn-v | challenge-v | **submit-v** | boast-v |

Table 3. Non-identified ESL synonyms for APSyn (window 2 and *N*=100), with their actual rank. The correct synonym is reported in bold.

In Table 4, we report the non-identified synonyms in the worst setting of *APSyn*, namely window 10 and *N*=1000. As it can be seen, most of those questions for which the synonym was not identified in the best setting (reported in *italics* in Table 4) are kept as errors also in the worst one. It is also interesting to notice that not only the number of errors increased, but also that the positions of the true synonyms in the error are lower than in the best setting. Even for questions that were already wrong in the best setting, the correct synonym was further penalized, losing a position (e.g. *refer-direct* and *yield-submit*). Finally, it is evident in Table 4 that several new non-identified synonyms are introduced, affecting negatively the overall performance.

To summarize, window 2 and *N* close to 100 are certainly the best parameters. Not only because they improve the overall accuracy, but also because, when making a mistake, they have a closer approximation to the correct answers. Namely the real synonym is typically ranked second rather than first, and the error is mostly due to sense ambiguity.

| Problem word | 1st | 2nd | 3rd | 4th |
|---|---|---|---|---|
| applause-n | shame-n | **approval-n** | fear-n | friend-n |
| approve-v | anger-v | boast-v | scorn-v | **support-v** |
| *grind-v* | *slice-v* | ***rub-v*** | *tap-v* | *hit-v* |
| *harvest-n* | *stem-n* | ***intake-n*** | *split-n* | *lump-n* |
| hinder-v | yield-v | assist-v | relieve-v | **block-v** |
| *limb-n* | *trunk-n* | ***branch-n*** | *bark-n* | *twig-n* |
| lump-n | limb-n | stem-n | trunk-n | **chunk-n** |
| mass-n | element-n | service-n | worship-n | **lump-n** |
| *passage-n* | *entrance-n* | ***hallway-n*** | *room-n* | *ticket-n* |
| ~~paste-n~~ | ~~syrup-n~~ | ~~**dough-n**~~ | ~~jelly-n~~ | ~~block-n~~ |
| *refer-v* | *carry-v* | *call-v* | *explain-v* | ***direct-v*** |
| scorn-v | avoid-v | enjoy-v | plan-v | **refuse-v** |
| *scrape-v* | *chop-v* | ***grate-v*** | *slice-v* | *mince-v* |
| *steep-j* | *rugged-j* | *bare-j* | ***sheer-j*** | *---* |
| ~~substance-n~~ | ~~level-n~~ | ~~**thing-n**~~ | ~~posture-n~~ | ~~score-n~~ |
| tap-v | knock-v | **drain-v** | rap-v | boil-v |
| *twist-v* | *clip-v* | *fasten-v* | *curl-v* | ***intertwine-v*** |
| verse-n | branch-n | twig-n | weed-n | **section-n** |
| *yield-v* | *challenge-v* | *scorn-v* | *boast-v* | ***submit-v*** |

Table 4. Non-identified ESL synonyms for APSyn (window 10 and *N*=1000), with their actual rank. The correct synonym is reported in bold. Errors that were present also in the best model are reported in italics, while errors that were present in the best, but absent in the worst, are reported with strikethrough.

The worst performance of *APSyn* is obtained for large window and large *N* (window 10 and *N*=1000). In the error analysis, we have seen that the worst model has generally bigger difficulty in classifying the synonym as somehow similar to the question word, often ranking it fourth.

## 7. Comparison with LMI-based *APSyn*

*APSyn* was introduced in Santus et al. (2016a) and tested on the ESL. In this paper, the top features were selected after ranking them by LMI instead of PPMI. LMI is less biased towards infrequent events than PPMI and it is defined as follows:

$$LMI(w, c) = P(w, c) \times PPMI(w, c)$$

where *w* is the target word, *c* is the given context, $P(w,c)$ is the probability of co-occurrence, as shown in the PPMI formula, above.

As mentioned above, the performance of the LMI-based *APSyn* on a window 5 DSM was worse than what reported with PPMI. However, its 58.33% accuracy was much above the *Vector Cosine*, which was instead blocked at 49.44%.

In Table 5 we show all the scores, recalculated with the LMI-based *APSyn*. Note that the recall in the models used for this paper is slightly higher, reaching 46 questions rather than 45, so the scores can be slightly different.

Despite results are worse than those obtained with PPMI, they are however relatively stable with reference to *N*, and almost always above the baseline. Only with window 10 and *N* close to 100 the performance is the equal to the *Vector Cosine*.

| APSyn | Win 2 | Win 3 | Win 5 | Win 10 |
|---|---|---|---|---|
| N=100 | 0.61 | 0.59 | 0.57[3] | 0.48 |
| N=200 | 0.65 | 0.61 | 0.48 | 0.48 |
| N=300 | 0.68 | 0.61 | 0.57 | 0.48 |
| N=400 | 0.66 | 0.64 | 0.50 | 0.52 |
| N=500 | 0.66 | 0.59 | 0.53 | 0.50 |
| N=600 | 0.66 | 0.59 | 0.53 | 0.50 |
| N=700 | 0.66 | 0.59 | 0.53 | 0.52 |
| N=800 | 0.66 | 0.57 | 0.53 | 0.52 |
| N=900 | 0.65 | 0.57 | 0.53 | 0.50 |
| N=1000 | 0.64 | 0.57 | 0.53[4] | 0.50 |
| Baselines | | | | |
| Cosine | 0.46 | 0.46 | **0.48** | **0.48** |
| Co-occ | 0.43 | 0.41 | 0.39 | 0.35 |

Table 5. Accuracy in the ESL test set for *APSyn* (100<*N*<1000), *Vector Cosine* and co-occurrence, in window 2, 3, 5 and 10 DSMs.

## 8. Conclusions

In this paper, we have described *APSyn*, a completely unsupervised measure based on the evaluation of the extent and the relevance of the intersection among the top ranked distributional features of target words. *APSyn* was tested on the ESL and TOEFL questions, outperforming the *Vector Cosine* and the *co-occurrence*, plus several lexicon-based and hybrid models. In particular, our results are above those reported in the literature for non-English US college applicants on the TOEFL test (64.50%).

Our experiments show that the intersection among the *N* most related contexts of the target words is in fact a reliable index of similarity. In our evaluations we have also mentioned the role of both the window size and *N*. *APSyn* performs better on smaller windows and with *N* close to 100. In fact, the larger the amount of considered contexts, the lower the ability of identifying similarity (exceptions to this consideration are minimal and can be appreciated only because of the limited size of the test sets). This also confirms our hypothesis that similar words share a significantly larger number of top mutually dependent contexts, but such intersection becomes less significant when not only the top contexts are considered, as rumor is introduced. Given that, it is important to notice that *APSyn* performance is quite stable, in respect to *N* variances.

*APSyn* has been recently used as one of the thirteen features of ROOT13, a random-forest based supervised system for the identification of hypernyms, co-hyponyms and unrelated words. In a *10-fold* evaluation on 9600 pairs extracted from EVALuation (Santus et al., 2015a), ROOT13 achieved 88.3% accuracy when the three classes were present, 93.4% for hypernyms-co-hyponyms discrimination, 92.3% for hypernyms-random discrimination, 97.3% for co-hyponyms-random (Santus et al., 2016b).

Possible improvements to the measure include changing the numerator to a more significant value, rather than simply using the constant 1. Moreover, it would be important to test the measure on optimized DSMs, where more parameters are investigated (e.g. stemming, dependency, SVD, etc.). Moreover, since ESL and TOEFL are small test sets, *APSyn* performance should be further explored on larger datasets, such as the Lenci/Benotto (Benotto, 2015), SimLex-999 (Hill et al., 2014) and EVALuation (Santus et al., 2015a).

## 9. Acknowledgements

This work is partially supported by HK PhD Fellowship Scheme under PF12-13656

## 10. Main References

---

[3] Not differently from Santus et al. (2016a), the LMI-based *APSyn* guessed 26.25 questions (24 full and 3 partial), but being the recall higher in the current DSM, this number has been divided by 46.

[4] Not differently from Santus et al. (2016a), the LMI-based *APSyn* guessed 24.25 questions (22 full and 3 partial), but being the recall higher in the current DSM, this number has been divided by 46.